\documentclass{article}
\usepackage{spconf,amsmath,graphicx,hyperref}
\usepackage[capitalize]{cleveref}             
\usepackage{enumitem}

\crefname{figure}{Figure}{Figures}  
\Crefname{figure}{Figure}{Figures}  
\crefname{table}{Table}{Tables}
\Crefname{table}{Table}{Tables}
\crefname{equation}{Equation}{Equations}
\Crefname{equation}{Equation}{Equations}

\title{ConfClip: Confidence-Weighted and Clipped Reward for Reinforcement Learning in LLMs}
\name{Author(s) Name(s)}
\address{Author Affiliation(s)}
\name{Bonan Zhang$^{\star \dagger}$ \qquad Zhongqi Chen$^{\dagger}$\qquad Bowen Song$^{\dagger}$ \qquad  Qinya Li$^{\star}$\qquad Fan Wu$^{\star}$ \qquad Guihai Chen$^{\star}$}
\address{
  $^{\star}$ Shanghai Jiao Tong University \\
  $^{\dagger}$ Ant Group
}

\begin{document}
%
\maketitle

\begin{abstract}
Reinforcement learning (RL) has become a standard paradigm for refining large language models (LLMs) beyond pre-training and instruction tuning. A prominent line of work is RL with verifiable rewards (RLVR), which leverages automatically verifiable outcomes (\textit{e.g.}, correctness or executability) to generate reward signals. While efficient, this framework faces two key limitations: First, its binary feedback is too sparse to capture the quality of the reasoning process. Second, its coarse-grained rewards potentially lead to vanishing gradients. Inspired by observations from human learning, we introduce a RL technique that integrates verifiable outcomes with the model’s own confidence estimates. This joint design enriches the reward signal, providing finer-grained feedback and implicitly supervising the reasoning process. Experimental results demonstrate that our proposed method enhances RL performance across multiple datasets and reduces token consumption during inference, while incurring negligible additional training cost. Moreover, it can be used as a plug-in module to enhance other state-of-the-art RL methods.
\end{abstract}
\begin{keywords}
Large Language Models, Reinforcement Learning, Fine-grained Reward
\end{keywords}
\section{INTRODUCTION}
\label{sec:intro}

Large Language Models (LLMs) have achieved remarkable progress in natural language processing, demonstrating strong generalization across diverse domains such as reasoning, code generation, and multi-step problem solving~\cite{brown2020language, achiam2023gpt, touvron2023llama}. Beyond large-scale pre-training and instruction tuning, reinforcement learning (RL) has emerged as a key technique for further aligning LLMs with desired behaviors~\cite{christiano2017deep, ouyang2022training}. By optimizing based on reward signals that encode correctness or preference, RL enables LLMs to refine their reasoning abilities and better satisfy human or task-specific requirements.

RL with verifiable rewards (RLVR)~\cite{guo2025deepseek} has recently emerged as a prominent framework for refining LLMs, which leverages automatically checkable signals such as exact correctness or executability. A notable instance is group relative policy optimization (GRPO)~\cite{shao2024deepseekmath} with RLVR. While attractive due to their scalability and efficiency, these methods suffer from inherent limitations. First, verifiable rewards are coarse-grained, as they only assess the final answer rather than the quality of the reasoning process. This binary feedback (correct or incorrect) fails to distinguish between nearly correct and entirely irrelevant responses, thus offering no insight into the soundness of the reasoning steps. Second, these binary rewards are sparse and prone to degeneration at two extremes in GRPO: when the task is overly simple, all outputs are correct, yielding no gradient; when the task is overly difficult, outputs are uniformly incorrect, again yielding no gradient and resulting in inefficient learning.

To overcome these challenges, we propose a novel reinforcement learning method, \textbf{ConfClip}, which integrates reward signals with model confidence. Our key insight, inspired by human learning, is that confidence provides complementary information to verifiable rewards: more confident correct or incorrect answers often reflect more accurate or flawed reasoning. By incorporating confidence into the reward, ConfClip produces a smoother reward and implicitly supervises the reasoning process: confidence captures fine-grained distinctions between partially correct and incorrect outputs, while verifiable correctness anchors learning to objective ground truth. Furthermore, we identify and analyze the instability that can arise when optimizing with confidence, introducing a simple yet effective solution. This ensures that ConfClip can be stably used as a plug-in module for other RL methods, improving performance with minimal additional computational overhead.

Our core contributions can be summarized as follows:
\begin{enumerate}[topsep=0pt,itemsep=2pt,parsep=0pt,partopsep=0pt]
    \item Motivated by the insight from human learning that confidence reflects the depth of understanding, we introduce a confidence-weighted reward mechanism that strengthens RLVR-based methods.
    \item Furthermore, we identify and explain the instability observed when directly using confidence-weighted rewards. We then propose a simple yet effective clip mechanism, enabling our method to serve as a plug-in module applicable to various RLVR methods.
    \item Our proposed method achieves consistent improvements across different model sizes and datasets. Experimental results further demonstrate the effectiveness of our design: ConfClip uses fewer tokens, maintains stable training, and directly improves the performance of state-of-the-art RLVR-based methods.

\end{enumerate}
\vspace{-5pt}
\section{Related Work}
\label{sec:format}
Large language models (LLMs) have recently been enhanced through reinforcement learning (RL), which offers advantages over supervised fine-tuning by directly optimizing non-differentiable objectives such as factuality, calibration, or reasoning accuracy. Among RL techniques, group relative policy optimization (GRPO) has gained attention for its efficiency and stability, enabling LLMs to leverage relative preference signals without requiring pairwise ranking models. In GRPO, the advantage is computed as follows:
\begin{equation}
\widehat{A}_i=\frac{r\left(x, y_i\right)-\operatorname{mean}\left(\left\{r\left(x, y_i\right)\right\}_{i=1}^G\right)}{\operatorname{std}\left(\left\{r\left(x, y_i\right)\right\}_{i=1}^G\right)},
\label{grpo}
\end{equation}
where $\widehat{A}_i$ denotes the advantage of the $i$-th response, 
$r\left(x, y_i\right)$ denotes the reward assigned to the $i$-th response $y_i$ from for a given problem $x$, $G$ denotes the response group size, $\text{mean}$ represents the mean reward within the group, 
and $\text{std}$ denotes the standard deviation.
A common practice in RL today is RL with verifiable rewards (RLVR). Specifically, in domains such as mathematics and code, rewards are assigned based on the correctness of the model's response. However, this approach results in sparse rewards that only supervise the final answer. Therefore, several studies have explored novel reward designs that rely on internal or self-derived signals. RENT~\cite{prabhudesai2025maximizing} introduces an entropy-minimization reward, encouraging models to generate more confident outputs and thereby improving reasoning accuracy in mathematical tasks. INTUITOR~\cite{zhao2025learning} leverages a self-certainty score, defined as the KL divergence from a uniform distribution, as an intrinsic reward. While these approaches yield certain improvements, depending only on the model’s confidence constrains its capacity for further learning. Rewarding Doubt~\cite{stangel2025rewarding} aligns model confidence with correctness by adopting a proper scoring rule as the reward. However, it relies on LLM-generated confidence, which, due to autoregressive decoding and the presence of hallucinations, can be unreliable~\cite{liu2025enhancing}.
\vspace{-5pt}
\section{METHODOLOGY}
\label{sec:pagestyle}
\subsection{Motivation}
\label{moti}

\begin{figure}[htb]
  \centering
  \centerline{\includegraphics[width=8cm]{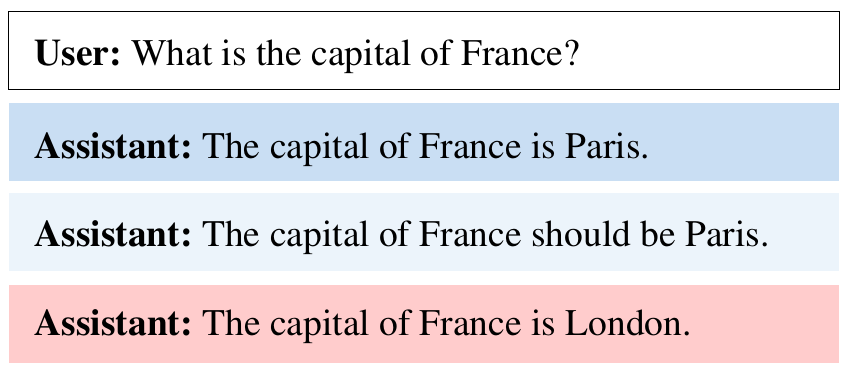}}
\vspace{-5pt}
\caption{A case illustrating our motivation. Blue shading indicates correct answers, while red shading indicates incorrect answers; darker colors correspond to higher confidence.}
\label{fig:toy1}
\vspace{-5pt}
\end{figure}
Our idea is inspired by human learning experience: if a student answers a question correctly without much confidence, it is likely to be a lucky guess, and thus the entire reasoning path should not be overly rewarded. Conversely, if a student is highly confident but gives an incorrect answer, this indicates a severe misunderstanding and should be penalized most heavily. The same holds in the opposite cases.

As shown in~\cref{fig:toy1}, compared with hesitant responses (Assistant Response 2, light blue), accurate and confident responses (Assistant Response 1, darker blue) are more appropriate. In contrast, confidently incorrect responses (Assistant Response 3, red) are precisely what should be avoided. Furthermore, we observe a phenomenon similar to earlier findings~\cite{wang2025beyond}: the model tends to exhibit lower confidence on tokens related to reasoning. Consequently, our method can reduce overthinking on simple problems while encouraging thoughtful reasoning to solve difficult tasks, which we empirically demonstrate in the experimental section.

\vspace{-7pt}
\subsection{Reweighting Rewards by Confidence}
Therefore, we propose to weight the correctness reward by the model’s confidence in its answer. For brevity, we denote the standard correctness reward $r(x, y_i)$ as $r_i$. Consequently, the confidence-weighted reward $\tilde{r}_i$ is given by:
\vspace{-5pt}
\begin{equation}
\tilde{r}_i = s_i r_i,
\end{equation}
where $s_i$ denotes the confidence coefficient. A heuristic approach is to use the probability of the final answer as the $s_i$. However, due to the autoregressive nature of language model generation, the answer is often effectively determined during the course of producing the response. This issue has become more pronounced with the rise of current reasoning models, making the probability of the final answer a poor proxy for the model’s actual confidence. 
Therefore, we use the probability of the entire generated sentence to represent the model’s confidence in its answer. Since the lengths of responses generated by language models can vary significantly, we normalize this confidence by the response length to prevent length bias, using it as our confidence coefficient $s_i$:
\vspace{-5pt}
\begin{equation}
s_i = \left({\pi_\theta(y_i \mid x)}\right)^{\tfrac{1}{|y_i|}}= \left(\prod_{t=1}^{|y_i|} 
   {\pi_\theta(y_{i,t}\mid x,y_{i,<t})}
   \right)^{\tfrac{1}{|y_i|}},
\end{equation}

where $\pi_\theta$ denotes the probability of the model with parameters $\theta$ generating the answer/token. 

\subsection{Penalizing Overconfidence}
Although we have weighted the reward scores, the commonly used 0/1 correctness reward $r_i$ gives all incorrect answers the same value $0$, eliminating differentiation. Therefore, we adopt a simple yet effective modification: changing the reward for incorrect answers from 0 to -1, so that overconfidence on wrong answers (\textit{e.g.}, \cref{fig:toy1}, Assistant Response 3, red) receives the maximum penalty:

\vspace{-5pt}
\begin{equation}
\tilde{r}_i =
\begin{cases}
s_i, & \text{if the answer is correct} \\[1mm]
-\,s_i. & \text{if the answer is incorrect}
\end{cases}
\vspace{-3pt}
\end{equation}

This modification also encourages the model to reason on difficult problems: when the task is difficult and the model’s responses are all incorrect, lower-confidence answers receive smaller negative rewards. This encourages the model to generate higher-entropy reasoning tokens, enhancing exploration during RL training and guiding the model to reason more effectively to solve the problem.
\vspace{-7pt}
\subsection{Stabilizing the Reward}
\label{Clip}
\vspace{-10pt}
\begin{figure}[htb]
  \centering
  \centerline{\includegraphics[width=8cm]{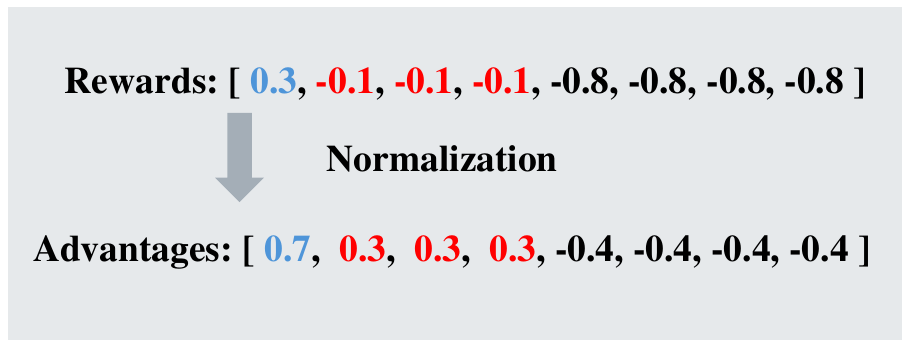}}
\vspace{-3pt}
\caption{A toy example illustrating how the update direction can be misleading. Due to the normalization operation in popular group-relative methods, on difficult problems, negative rewards may produce positive gradients, causing the model to learn in the direction of low-confidence incorrect answers.}
\label{fig:toy}
\end{figure}

Although current algorithms have achieved significant progress on GRPO, we observed that they are sensitive to hyperparameters and occasionally experience training collapse. Upon analysis, we found that this phenomenon originates from situations where the dataset is relatively difficult for the base model, causing the majority of answers in a group to be incorrect in the early stages of training. For example, as shown in~\cref{fig:toy}, due to the presence of normalization, relatively low negative rewards can become positive advantages. When the problem is difficult and correct answers are scarce, the total advantage of incorrect answers may even exceed that of a few correct ones. Since the gradient is proportional to the advantage, updates in the direction of incorrect answers can outweigh those in the direction of correct answers. As a result, the model tends to always produce incorrect but low-confidence answers to minimize the penalty.
Therefore, we constrain the model's reward within a certain range to prevent excessively large updates in the direction of incorrect answers:
\begin{equation}
\tilde{r}_i =
\begin{cases}
\mathrm{clip}(s_i, 1-\epsilon, 1) & \text{if the answer is correct} \\[1mm]
\,\mathrm{clip}(-s_i, -1, \epsilon-1), & \text{if the answer is incorrect}
\end{cases}
\end{equation}
where $\epsilon$ is a hyperparameter controlling the clipping range.
\vspace{-5pt}

\section{EXPERIMENTS}

\subsection{Training Setup}
Following the prior work~\cite{zhao2025learning}, we adopt Qwen2.5-3B~\cite{qwen2.5} as our base model and employ GRPO~\cite{shao2024deepseekmath} as the baseline algorithm. Training is conducted for one epoch on the MATH~\cite{hendrycksmath2021} dataset, utilizing an on-policy setup with a batch size of 128 for both rollout and gradient updates. Each question in the rollout generates seven candidate answers, and a KL divergence penalty of 0.005 is applied to the loss function. For a fair comparison, ConfClip uses the same hyperparameters as the baseline, with the hyperparameter of clipping set to 0.2, following standard practices. The training process is implemented within the verl~\cite{sheng2024hybridflow} framework.

\subsection{Evaluation} We evaluate our method on the math datasets MATH~\cite{hendrycksmath2021}, GSM8K~\cite{cobbe2021gsm8k}, and AIME24~\cite{li2024numinamath}, and use the multi-domain dataset MMLU-Pro~\cite{wang2024mmlu} to assess generalization ability. For the math datasets, we consistently use the same chat templates as in training. For MMLU-Pro, we follow the 5-shot testing templates from its original paper~\cite{wang2024mmlu}. On MATH, GSM8K, and MMLU-Pro, we report the accuracy obtained using greedy decoding. To ensure stable results on the AIME dataset, which contains relatively few questions, we employ a temperature of 1.0 and perform 32 inference runs for each question. We report both the proportion of questions that were correctly solved in at least one run and the overall average success rate. 
\begin{table}[b]
\setlength{\tabcolsep}{5pt}
\centering
\vspace{-15pt}
\caption{Test accuracy (\%) of the methods.}
\label{tab:qwen3}
{\fontsize{9}{13}\selectfont
\begin{tabular}{c|ccc|c}
\hline
Qwen2.5-3B & MATH       & GSM8K          & AIME24              & MMLU-Pro       \\ \hline
Base       & 30.24          & 68.16          & 10.00/0.82          & 37.55          \\ \hline
GRPO       & 63.07          & 78.39          & 13.33/1.77          & 39.43          \\ \hline
\textbf{ConfClip} & \textbf{64.34} & \textbf{79.23} & \textbf{20.00/5.21} & \textbf{39.84}         \\ \hline
\end{tabular}}
\end{table}
\vspace{-5pt}
\subsection{Results}
The results in~\cref{tab:qwen3} demonstrate that our method not only delivers substantial improvements on in-domain tasks but also achieves performance on the out-of-domain MMLU-Pro dataset that matches or surpasses GRPO. This indicates that the performance gains of our method do not stem from potential overfitting.
\begin{table}[t]
\setlength{\tabcolsep}{3pt}
\centering
\caption{Test accuracy (\%) of the methods.}
\label{tab:qwen7}
{\fontsize{9}{13}\selectfont
\begin{tabular}{c|ccc|c}
\hline
Qwen2.5-7B & MATH       & GSM8K          & AIME24              & MMLU-Pro       \\ \hline
Base       & 32.44         & 71.04          & 16.67/2.29          & 48.78          \\ \hline
GRPO       & 71.49          & 82.26          & 20.00/6.25          & \textbf{50.34} \\
\textbf{ConfClip} & \textbf{72.36} & \textbf{85.22} & \textbf{20.00/6.53} & 50.17          \\ \hline
GSPO       & 71.94          & 84.00          & \textbf{23.33}/4.27          & 50.37          \\
\textbf{GSPO+ConfClip} & \textbf{72.73} & \textbf{84.88} & 16.67/\textbf{5.41}          & \textbf{50.71} \\ \hline
\end{tabular}}
\end{table}

\begin{figure}[t]
    \centering
    \begin{minipage}[t]{0.47\columnwidth}
        \centering
        \includegraphics[width=\textwidth]{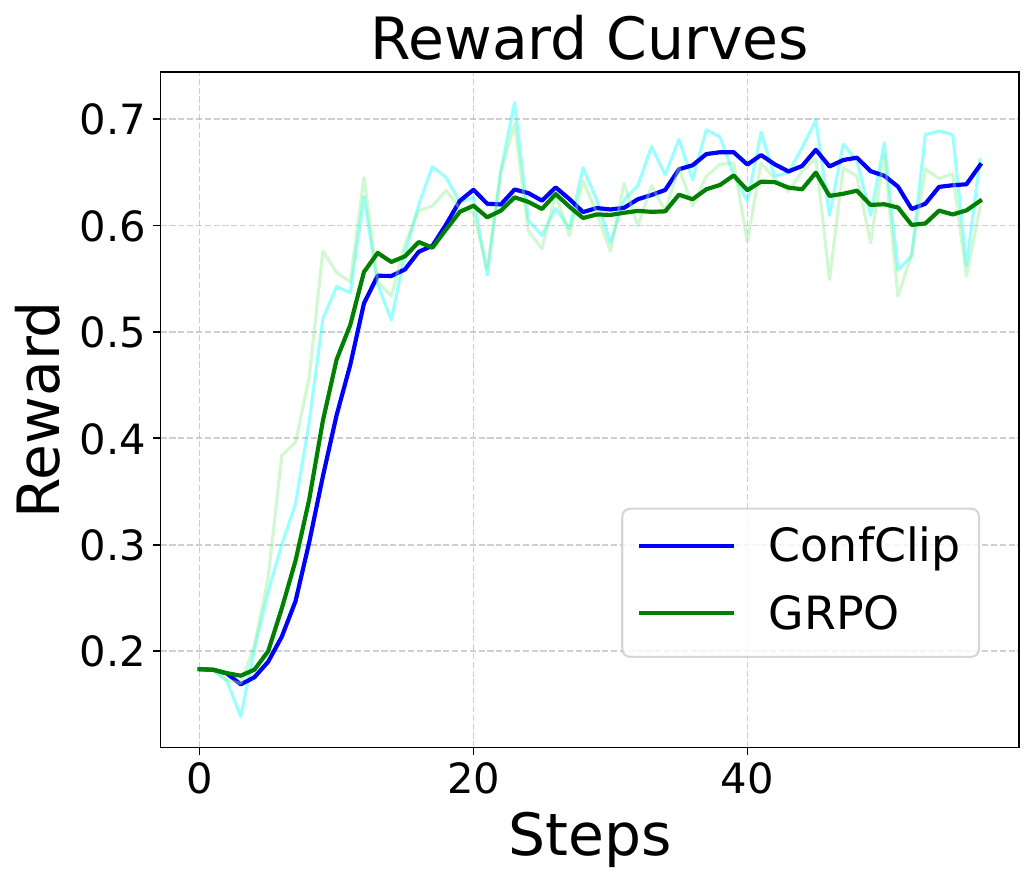}
    \end{minipage}%
    \hfill
    \begin{minipage}[t]{0.47\columnwidth}
        \centering
        \includegraphics[width=\textwidth]{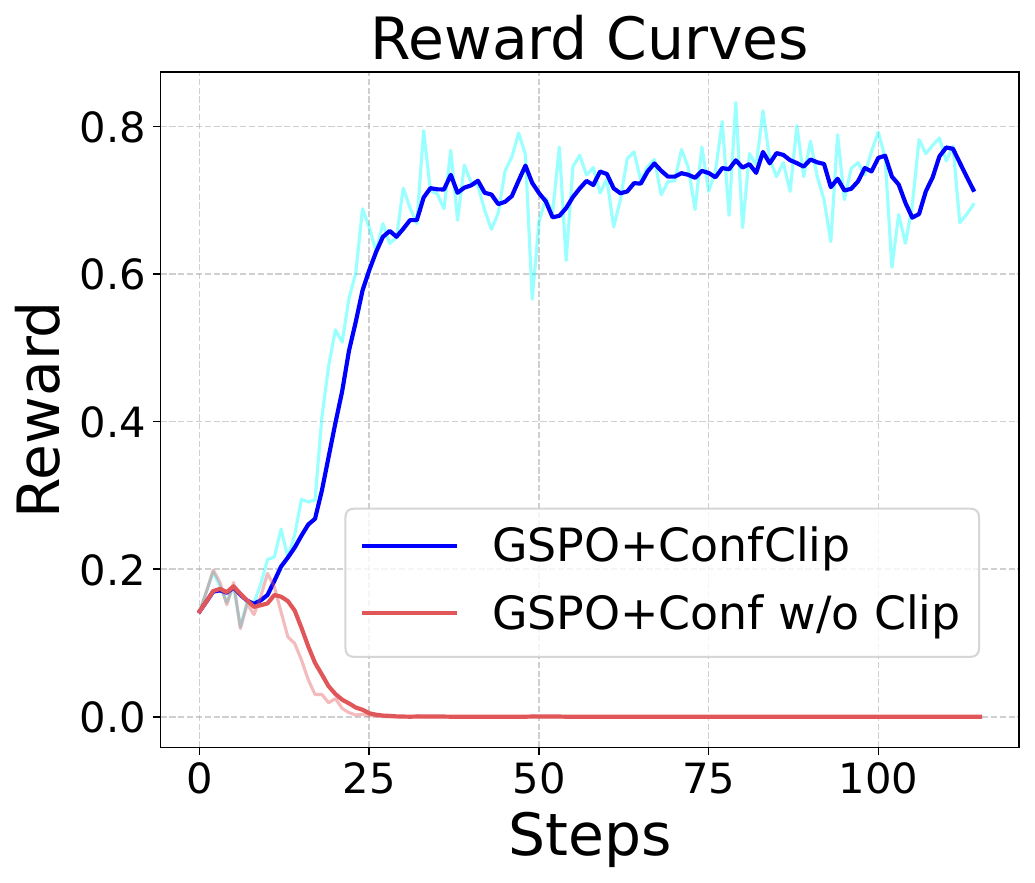}
    \end{minipage}
    \caption{Visualization of the correctness reward curves.}
    \vspace{-10pt}
    \label{fig:curve}
\end{figure}

\textbf{Learning Dynamics.} Furthermore, our method exhibits training stability comparable to the baseline. As shown in~\cref{fig:curve} (left), ConfClip gradually discovers high-confidence answers, which drive a stable increase in rewards beyond the baseline. While the correctness reward for a wrong answer is -1 in ConfClip, it is plotted as 0 in the figures to ensure a visual comparison with the baseline. The light and dark-colored curves in the figures represent the raw data and its 5-step moving average, respectively.
\vspace{3pt}

\begin{figure}[t]
	\centering
	\includegraphics[width=0.8\linewidth]{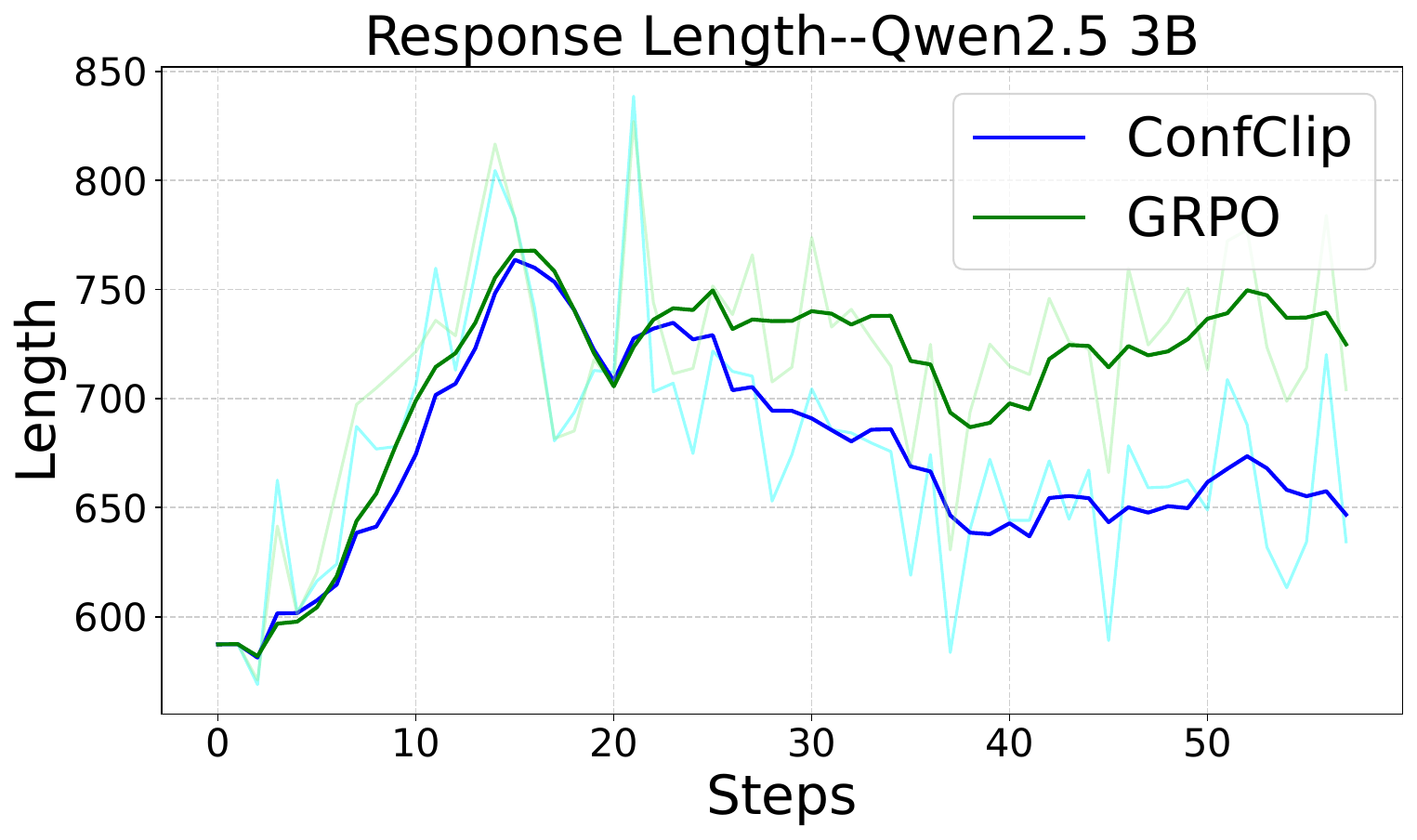}
	\caption{Average response length of the models during training.}
	\label{fig:length}
\end{figure}

\textbf{Suppressing Overthinking.}
 As discussed in~\cref{moti}, on problems that the model can already solve, our method encourages more confident responses, thereby suppressing overthinking. This reduces the output length and saves tokens. As shown in ~\cref{fig:length}, during the early stage of exploration, our method produces outputs of similar length to the baseline. However, in the later stages of training, our method achieves higher accuracy while consuming fewer tokens.
\vspace{3pt}

\textbf{Scaling to Larger Models.}
We also evaluate ConfClip using the Qwen2.5-7B base model~\cite{qwen2.5} to evaluate its scalability, with the rollout size increased to 14. As shown in~\cref{tab:qwen7}, our method remains effective when applied to larger-scale model. On GSM8K, it even exhibits greater performance gain, which we attribute to the larger models’ ability to provide more accurate confidence estimates for their responses. This highlights the potential of our method.
\vspace{3pt}

\textbf{Compatibility to SOTA RL Methods.}
Since our method only modifies the reward, making it a plug-in module that can be integrated with various LLM RL methods, we additionally conduct experiments using the latest SOTA algorithm GSPO~\cite{zheng2025group} as the baseline, keeping the same hyperparameters as in GRPO.
\vspace{-7pt}
\subsection{Ablation Study}
To further investigate the effect of clip operation, we conduct an ablation study to demonstrate its importance.  We use GSPO as the baseline in our ablation study.~\cref{fig:curve} (right) visualizes the mean reward during training, showing that the model rapidly collapses into a chaotic state without clip operation, producing only incorrect answers after a few steps.

\begin{figure}[t]
	\centering
	\includegraphics[width=0.8\linewidth]{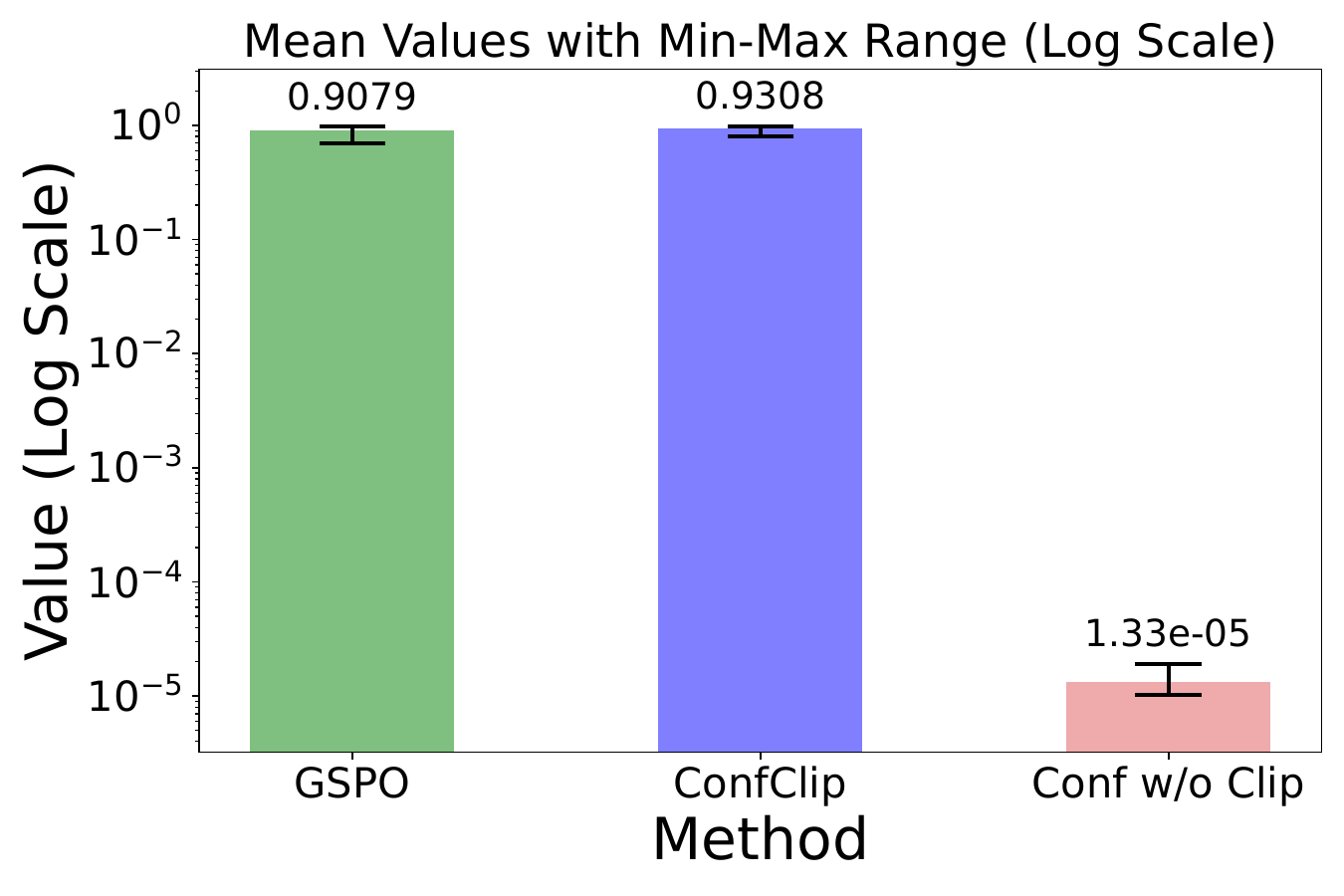}
    \vspace{-5pt}
	\caption{Confidence of the final-step response.}
    \vspace{-5pt}
	\label{fig:conf}
\end{figure}

Moreover, we empirically validate the conclusions presented in~\cref{Clip} by visualizing the output sentence probabilities at the final step for GSPO, GSPO+ConfClip, and GSPO+Conf without Clip. As shown in~\cref{fig:conf}, it can be observed that GSPO+ConfClip exhibits confidence levels similar to GSPO, whereas the confidence in GSPO+Conf without Clip is several orders of magnitude lower. Combined with the reward collapse observed in~\cref{fig:curve} (right), these results strongly support our conclusion: due to the presence of normalization, the model may be inclined to produce low-confidence incorrect answers to reduce the effective penalty.

\vspace{-3pt}
\section{CONCLUSION}
\vspace{-2pt}
In this work, we propose ConfClip, which reweights the reward based on the model’s confidence in its outputs. Inspired by human learning, where more confident correct or incorrect answers often reflect the more reasonable or unreasonable thinking, our approach implicitly scores the reasoning process. Furthermore, we identify and explain the training instability that arises when using confidence-weighted rewards and introduce a simple yet effective solution. Experiments on datasets with varying difficulty levels further validate the effectiveness of our design.

\vfill\pagebreak
\bibliographystyle{IEEEbib}
\bibliography{main,refs}

\end{document}